\documentclass[runningheads]{llncs}

\usepackage{eccv}

\usepackage{eccvabbrv}
\usepackage{tcolorbox}
\tcbuselibrary{breakable}
\usepackage{listings}
\usepackage{xcolor}

\usepackage{booktabs}
\usepackage{multirow}
\usepackage{subcaption}
\usepackage{adjustbox}

\newcounter{ablationpart}[table]

\lstdefinestyle{promptstyle}{
    basicstyle=\ttfamily\scriptsize,
    breaklines=true,
    columns=fullflexible,
    frame=none,
    aboveskip=2pt,
    belowskip=2pt
}

\lstdefinestyle{promptstyle}{
    basicstyle=\ttfamily\scriptsize,
    breaklines=true,
    columns=fullflexible,
    frame=none,
    aboveskip=2pt,
    belowskip=2pt
}

\tcbset{
promptbox/.style={
    colback=gray!6,
    colframe=black,
    boxrule=0.5pt,
    arc=2pt,
    boxsep=2pt,
    left=3pt,
    right=3pt,
    top=3pt,
    bottom=3pt,
    breakable
}}

\usepackage{graphicx}
\usepackage{booktabs}
\usepackage{multirow}
\usepackage{enumitem}

\usepackage[accsupp]{axessibility}  

\usepackage{hyperref}

\usepackage{orcidlink}
\newcommand{\equalcontrib}{\textsuperscript{*}}
\newcommand{\corrauthor}{\textsuperscript{\ensuremath{\dagger}}}
\newcommand{\internship}{\textsuperscript{\ensuremath{\ddagger}}}

\usepackage{graphicx}
\usepackage{subcaption}

\makeatletter
\newcommand{\inlineand}{%
  \stepcounter{@inst}%
  \qquad
  \noindent$^{\the@inst}$\enspace\ignorespaces
}
\makeatother

\begin{document}

\title{HAT-4D: Lifting Monocular Video for 4D Multi-Object Interactions via Human-Agent Collaboration}

\titlerunning{HAT-4D}

\author{Jiaxin Li\inst{1,2} \equalcontrib  \orcidlink{0009-0009-2422-300X}\and
Yuxiang Wu\inst{1} \equalcontrib \orcidlink{0009-0003-8827-7839} \and
Zhenkai Zhang\inst{1}\orcidlink{0009-0007-4324-9075} \and
Xinrui Shi\inst{1}\orcidlink{0009-0005-7449-2153} \and
Haoyuan Wang\inst{1}\orcidlink{0009-0009-9338-9556} \and
Yichen Zhao\inst{1} \internship \orcidlink{0009-0004-0185-0143} \and
Su Linxiang\inst{1} \internship \orcidlink{0009-0009-3415-7931}\and
Chenyang Yu\inst{1}\orcidlink{0009-0009-3958-2590}\and
Mingyu Zhang\inst{1,2}\orcidlink{0009-0009-4174-9601}\and
Yifan Ding\inst{2}\orcidlink{0009-0007-1369-088X}\and
Boran Wen\inst{1,2}\orcidlink{0009-0000-8189-5472}\and
Li Zhang\inst{3}\orcidlink{0000-0003-1610-6056}\and
Ruiyang Liu\inst{4}\orcidlink{0000-0003-0075-6230}\and
Yong-Lu Li\inst{1,2} \corrauthor \orcidlink{0000-0003-0478-0692}}

\authorrunning{J. Li et al.}

\institute{
Shanghai Jiao Tong University, China
\inlineand Shanghai Innovation Institute, China
\and
University of Science and Technology of China, China
\inlineand Math Magic, China\\
\email{\{li\_jiaxin, vx.limo, yonglu\_li\}@sjtu.edu.cn}
}

\maketitle

\begingroup
\renewcommand{\thefootnote}{*}
\footnotetext[1]{Equal contribution.}

\renewcommand{\thefootnote}{\ensuremath{\dagger}}
\footnotetext[2]{Corresponding author.}

\renewcommand{\thefootnote}{\ensuremath{\ddagger}}
\footnotetext[3]{Conducted during an internship at Shanghai Jiao Tong University.}
\endgroup

\begin{abstract}
Extracting dynamic 4D object interactions from massive, in-the-wild monocular videos offers a highly efficient data collection pathway for scaling Embodied AI and training VLAs. 
However, existing monocular 4D reconstruction methods primarily focus on isolated objects, often failing under the severe occlusions and complex dynamics inherent in multi-object interactions. 
To bridge this gap, we propose \textbf{HAT-4D}, the first \textit{agentic framework} designed to reconstruct the 3D geometry, temporal dynamics, and physical interactions of multiple objects from a single video. 
By integrating VLMs with a multi-level \textit{human-in-the-loop} feedback mechanism, HAT-4D efficiently resolves depth ambiguities and interaction-induced occlusions during 3D generation and 4D propagation, yielding physically plausible assets without relying on expensive multi-camera rigs. 
As a scalable data engine, HAT-4D facilitates the creation of \textbf{MVOIK-4D}, an open-world benchmark for monocular 4D interaction reconstruction, accompanied by a novel multi-dimensional evaluation protocol focused on physical plausibility and temporal consistency.
Extensive experiments demonstrate that HAT-4D
achieves SOTA performance on most evaluation metrics,
while maintaining competitive semantic alignment.
Ablation studies show that introducing a small amount of
human feedback improves interaction reconstruction.
Moreover, the data produced by HAT-4D effectively improves baseline performance when used for finetuning.
Our data and code are available at \href{https://lijiaxin0111.github.io/HAT4D/}{project webpage}.
\end{abstract}
\section{Introduction}
\label{sec:intro}

\begin{figure*}[t]
    \centering
    \includegraphics[width=\textwidth]{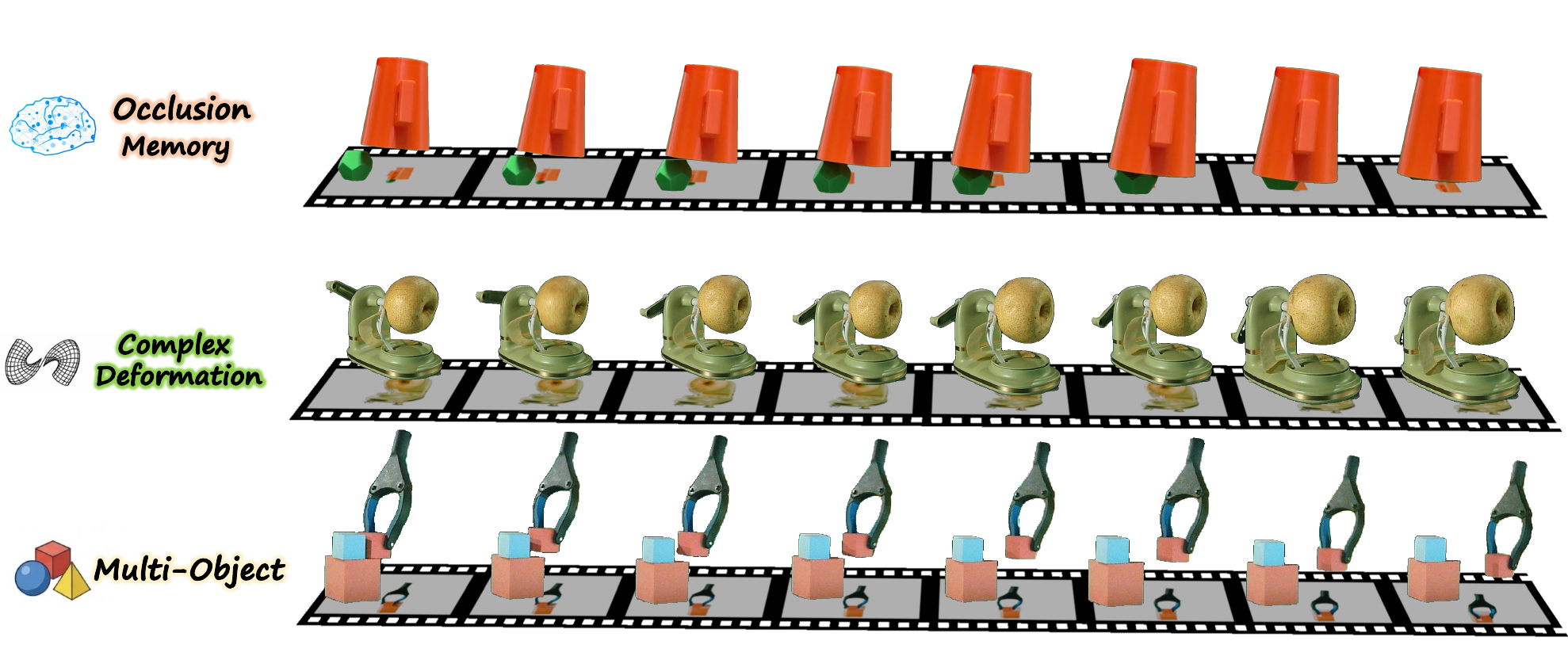}
    \caption{
        \textbf{Examples of object interaction sequences from the MVOIK-4D dataset.}
        The dataset captures diverse real-world interaction scenarios with rich
        spatio-temporal dynamics. It includes challenging cases such as
        \textbf{occlusion memory}, where objects disappear and reappear across
        time, \textbf{complex deformation} caused by physical manipulation,
        and \textbf{multi-object interactions} involving coordinated motion
        between multiple objects. These scenarios require models to reason
        about geometry, dynamics, and long-term temporal consistency simultaneously.
    }
    \label{fig:teaser}
\end{figure*}

The ability to understand and reconstruct physically plausible object interactions within a spatio-temporally coherent 4D frame is essential for numerous downstream applications, such as populating immersive virtual environments~\cite{li2025genhoigeneralizingtextdriven4d,ji2025immersivehumanxinteractionrealtime}, and training robotic perception\cite{chen2025m}, understanding\cite{zhang2026ipr,li2023beyond}, and manipulation~\cite{hou20254dvisualpretrainingrobot,zeng2024learningmanipulationpredictinginteraction,wang2026great}. 
Since monocular video serves as the most abundant and accessible source of real-world object interactions, lifting these 2D observations into dynamic 4D multi-object interactions has emerged as a central objective for the research \cite{l4gm,fb4d,gvfdiffusion}. However, this reconstruction is a fundamentally ill-posed problem, requiring the disentanglement of complex, evolving spatial relationships, the resolution of heavy mutual occlusions, and the maintenance of strict spatio-temporal consistency across the sequence.

Existing approaches to acquiring 4D object interaction data generally fall into two extremes. On the one hand, hardware-intensive approaches~\cite{lu2024diva,chen20254dslomo} rely on massive, synchronized multi-camera rigs. While they capture occlusion-free objects with high-quality dynamics, these systems are prohibitively expensive, strictly confined to controlled studio environments, and fundamentally unscalable for open-world data collection.
On the other hand, recent generative approaches attempt to bypass these hardware constraints by leveraging diffusion-based priors or synthetic data to either animating static 3D collections~\cite{objaverse,objaversexl,hy3dbench} towards 4D settings~\cite{chen2026motion,mou2025dimo} or to reconstruct the spatial frame from monocular video inputs~\cite{4d-lrm,l4gm,gvfdiffusion,stag4d,fb4d}. However, current generative models primarily focus on isolated, stylized assets, such as gaming or cartoon characters. When confronted with the complex and diverse object interactions present in the real world, they suffer from severe domain gaps that lead to unstable generation quality, manifests as implausible physical-deformations, floating interactions, and noticeable temporal jittering. 

To break the deadlock between the unscalable cost of multi-view capture and the unreliable quality of generative models, we propose \textbf{HAT-4D}, a low-cost, \textit{human-in-the-loop (HITL)} agentic framework for monocular 4D object interaction generation. Given an in-the-wild monocular video, HAT-4D first leverages Vision-Language Model (VLM) agents to organize the visual sequence into a structured \textbf{Interaction Knowledge Graph (IKG)}. Serving as the core causality engine of our framework, the IKG efficiently encodes long-term physical changes and the underlying interaction cues that drive them. Guided by IKG, HAT-4D orchestrates a suite of specialized agents: it couples \textit{3D object generation and composition skills} to recover the precise geometry and spatial alignment of interacting entities, and subsequently applies \textit{4D propagation skills} to reconstruct their evolving dynamics. To explicitly resolve the severe depth and occlusion ambiguities inherent to monocular video, we integrate high-level human knowledge into the generation loop via a novel HITL collaborative scheme, empowering users to actively guide and refine the 3D generation, spatial composition, and dynamic 4D propagation processes. 

Leveraging HAT-4D as a highly scalable data engine, we construct \textbf{MVOIK-4D (Multi-View Object Interaction 4D Knowledge)} benchmark encompassing  77 tasks across 112 distinct interaction scenarios, alongside a novel multi-dimensional evaluation protocol designed to rigorously assess physical plausibility and interaction stability involving deformation realism, interaction consistency, temporal smoothness, cross-view and long-term memory preservation. Extensive experiments validate that HAT-4D consistently outperforms existing monocular 4D baselines in modeling complex multi-object interactions, paving the way for downstream Embodied AI research by supplying high-fidelity, scalable physical priors.

In summary, our main contributions are: 
(1) We propose \textbf{HAT-4D}, an innovative human-in-the-loop multi-agent system designed to reconstruct physically plausible and temporally coherent 4D multi-object interactions directly from monocular videos.
(2) We introduce \textbf{Interaction Knowledge Graph (IKG)} as the core causality engine of our framework. By explicitly encoding long-term physical changes, the IKG guides our specialized 3D generation and 4D propagation agents to resolve severe depth ambiguities and mutual occlusions.
(3) We establish \textbf{MVOIK-4D}, a comprehensive benchmark encompassing a wide range of interaction scenarios, coupled with a novel multi-dimensional evaluation protocol to assess physical plausibility and temporal consistency.

\section{Related Work}

\subsection{Object Interaction Knowledge Understanding  }

Early studies on object interaction understanding mainly focus on 
image-level reasoning, such as object relation detection\cite{krishna2016visualgenomeconnectinglanguage,li2022sgtrendtoendscenegraph,liu2025interacted} and object affordance detection.\cite{do2018affordancenetendtoenddeeplearning,chen2023affordancegroundingdemonstrationvideo} 
With the development of vision-language models (VLMs), 
Recent works extend this direction to long videos, 
enabling richer interaction understanding through temporal reasoning \cite{cheng2024videollama2advancingspatialtemporal,wang2024internvideo2scalingfoundationmodels}.

With the progress of 3D reconstruction \cite{liu2023zero1to3zeroshotimage3d,hong2024lrmlargereconstructionmodel,wang2025vggtvisualgeometrygrounded} and dynamic scene modeling \cite{wu20244dgaussiansplattingrealtime,zhang2025efficientlyreconstructingdynamicscenes}, object interaction understanding in the 4D domain has begun to attract attention. 
In 4D settings, interactions involve complex spatio-temporal dynamics 
beyond static semantic relationships. 
However, collecting large-scale real-world 4D interaction data is difficult. 
As a result, existing datasets are still largely limited to synthetic environments.

In this work, we study real-world 4D object interaction understanding with an agent-driven framework. 
We construct a multi-view object interaction dataset with rich annotations covering diverse interaction scenarios. 
To explicitly represent complex interactions in dynamic scenes, 
we introduce an \textbf{interaction knowledge graph} that models object relationships, 
interaction categories, and deformation states over time. 
This structured representation captures the underlying spatio-temporal interaction dynamics 
and provides guidance for reconstructing 4D object interactions from monocular videos. 

\subsection{Monocular Video-based 4D Generation}

With the development of the video diffusion model~\cite{stablediffusion,svd,sv3d} and a large-scale dynamic 3D object dataset~\cite{objaverse,objaversexl,vividzoo,texverse}, 4D content generation attracts more and more attention.
Recent research in 4D generation can be broadly divided into two paradigms.
The first paradigm is optimization-based, leveraging pretrained image or video diffusion models through Score Distillation Sampling (SDS) to extract 4D features~\cite{consistent4d,stag4d,birth,fb4d}. 
However,  Score Distillation Sampling (SDS) often suffers from the Janus problem when generating the complex object.
The second paradigm follows a dataset-driven, end-to-end approach that relies on large-scale 3D or 4D datasets~\cite{sv4d,l4gm,gvfdiffusion,v2m4}.
Models such as SV4D~\cite{sv4d}, L4GM~\cite{l4gm}, and GVF-Diffusion~\cite{gvfdiffusion} extend high-fidelity 3D generation architectures by introducing spatio-temporal cross-attention layers, enabling end-to-end dynamic scene generation. 
However, these models are typically trained on large-scale synthetic datasets. 
When applied to real-world scenes, their performance often degrades due to the significant domain gap.

To address these challenges, we propose \textbf{HAT-4D}, an agent-driven framework that integrates multi-level human-in-the-loop feedback into the 4D generation process. 
By introducing human corrections during generation, the framework effectively reduces error accumulation and improves the physical plausibility of generated interactions. 
It also mitigates memory degradation when objects undergo heavy occlusions or reappear after long temporal gaps.  
Furthermore, HAT-4D serves as an efficient 4D data engine to construct the \textbf{MVOIK-4D} benchmark, alleviating the lack of real-world 4D object interaction data.

\subsection{Human-in-the-loop Tool in Visual Tasks}

Human-in-the-loop (HITL) methods have long been closely related to data construction and the scarcity of annotated data. Early studies mainly explored HITL through active learning~\cite{settles2009active}, where models identify uncertain samples and request human annotations, and such strategies have been widely applied in object detection~\cite{yu2022consistencybasedactivelearningobject} and medical imaging tasks~\cite{Wang_2019,Budd_2021}.

With the development of deep learning, HITL has evolved into a data engine paradigm. For example, Segment Anything~\cite{kirillov2023segment} introduces a human–model data engine that iteratively expands large-scale segmentation datasets through model prediction and human correction, enabling strong segmentation performance in open-world scenarios.
Recent works extend this idea to the 3D domain~\cite{cen2024segment3dradiancefields,wang2019latteacceleratinglidarpoint}, where model predictions are refined through human evaluation or expert correction to construct large-scale 3D training data and alleviate the scarcity of real-world 3D datasets.

In this work, we further extend this paradigm to dynamic 4D reconstruction. Leveraging the few-shot capabilities of vision–language models (VLMs), we propose HAT-4D, a human–agent collaborative framework that addresses the scarcity of real-world 4D interaction data. Our agent performs automatic repair during generation, while human feedback is introduced at sparse keyframes to correct accumulated errors, enabling reliable reconstruction of complex object interactions.
\section{Method}

\begin{figure*}[!t]
    \centering
     \begin{subfigure}{\textwidth} 
        \centering
        \includegraphics[width=\linewidth]{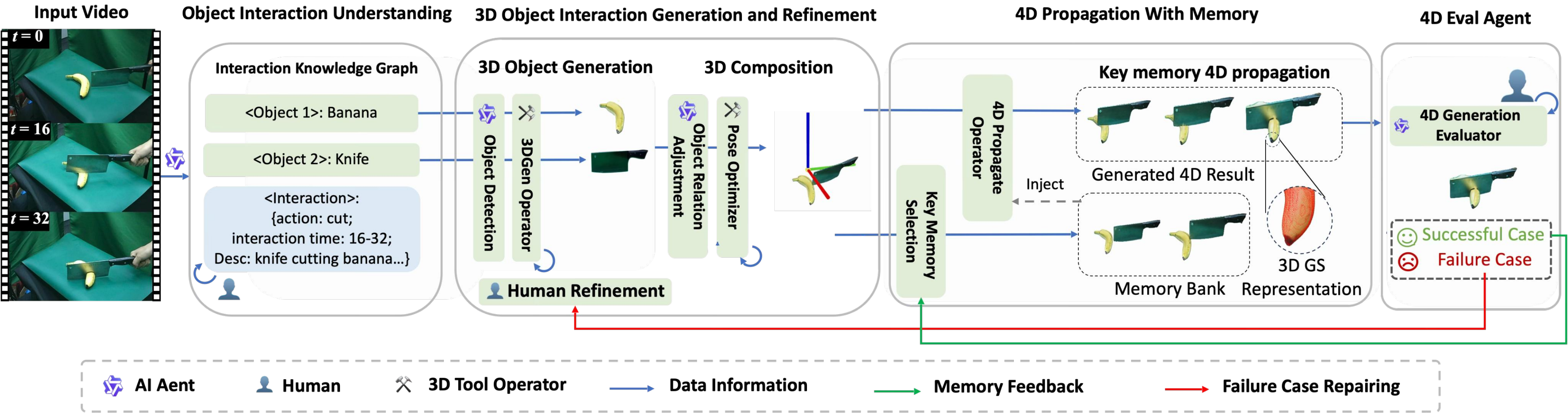} 
       \end{subfigure}
    
\caption{\textbf{Overview of HAT-4D.}
The framework extracts an \textbf{interaction knowledge graph} from a monocular video, reconstructs and composes the interacting objects, and generates subsequent 3D states through \textbf{memory-guided }4D propagation.
The evaluation agent updates the memory with successful results and feeds failed cases back for repair, while human users can refine intermediate outputs throughout the pipeline.} \label{fig:method_pipeline_camerady}

\end{figure*}

Our objective is to reconstruct 4D multi-object interactions from monocular videos, a task hindered by severe depth ambiguities, heavy occlusions, and evolving topological states. 
To overcome these challenges and mitigate error accumulation, we propose \textbf{HAT-4D}, an agentic framework (Fig,~\ref{fig:method_pipeline_camerady}) driven by an explicit \textbf{Interaction Knowledge Graph (IKG)}. Within this framework, interacting entities and their physical deformations are reconstructed as 4D Gaussian Splats\cite{wu20244dgaussiansplattingrealtime} and organized into critical interaction event segments.
Guided by the IKG, our specialized agents ensure physically plausible 3D generation and temporally consistent 4D propagation, while multi-level \textbf{human-in-the-loop (HITL)} collaboration enables precise geometric refinement and semantic editing. The remainder of this section is organized as follows: Sec.~\ref{sec:ikg} details the formulation of the \textbf{IKG}; Sec.~\ref{sec:hat4d} describes the \textbf{HAT-4D} agentic system, encompassing 3D generation, spatial composition, and memory-augmented 4D propagation skills; and Sec.~\ref{sec:hitl} introduces our multi-level \textbf{HITL} collaborative mechanisms.

\subsection{Interaction Knowledge Graph Formulation}
\label{sec:ikg}

\begin{figure*}[!t]
    \centering
     \begin{subfigure}{\textwidth} 
        \centering
        \includegraphics[width=\linewidth]{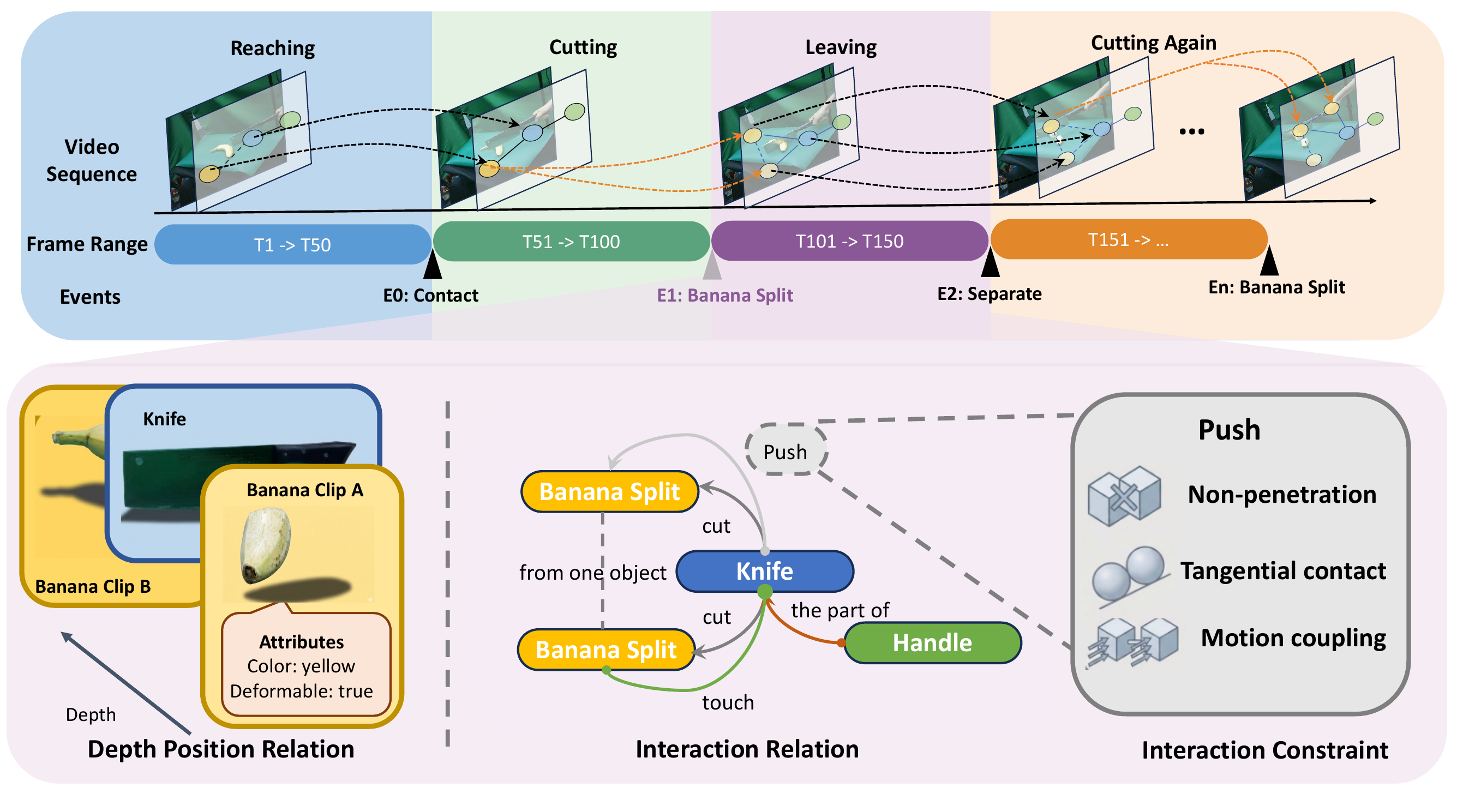} 
       \end{subfigure}
    
   \caption{
\textbf{Overview of the Interaction Knowledge Graph (IKG) formulation.}
The IKG represents physical interactions as a dynamic, temporally conditioned directed graph.
\textbf{Top:}
The video sequence is partitioned into event segments ($\mathcal{E}$) according to state transitions 
(e.g., \textit{E0: Contact}, \textit{E1: Banana Split}). 
Each segment models interaction phases such as \textit{Reaching}, \textit{Cutting}, and \textit{Leaving}.
\textbf{Bottom Left:}
Depth cues and physical attributes (e.g., color, deformability) are extracted for object entities ($\mathcal{O}$).
\textbf{Bottom Center \& Right:}
Spatio-temporal relations ($\mathcal{R}$) and interaction constraints 
(e.g., non-penetration, motion coupling) encode semantic and physical dependencies 
between objects (e.g., \textit{Knife} and \textit{Banana}), guiding downstream 4D generation.
} \label{fig:OIK}
    
\end{figure*}

The fundamental novelty of HAT-4D lies in leveraging the understanding of physical interactions to guide the generation of visual assets. Given an input monocular video, we first prompt Qwen3-VL~\cite{bai2025qwen3vltechnicalreport} to digest the entire sequence and formalize it into an IKG (Fig.~\ref{fig:OIK}). Mathematically, the IKG is defined as a dynamic, temporally conditioned directed graph $\mathcal{G}=(\mathcal{O},\mathcal{E},\mathcal{R})$ that tracks object entities $\mathcal{O}$, interaction event segments $\mathcal{E}$ over the video duration $T$, and spatial and semantic relationships $\mathcal{R}$ between those entities within each segment.

\textbf{Object Entities.} $\mathcal{O}=\{o_1,o_2,...,o_n\}$ represents the set of distinct interactable objects in the scene (e.g., banana and knife). Each object $o_i$ encapsulates semantic metadata and physical attributes that hints the downstream generation and validation, including colors, textures, deformability, and affordance.   

\textbf{Interaction Event Segments.} $\mathcal{E}$. The temporal structure of the interaction is represented by a set of event segments $\mathcal{E}=\{e_1,e_2,...,e_m\}$, partitioned by state changes, such as the severing of an object, and occlusion relation shifts. Each segment $e_i\in\mathcal{E}$ is defined by its temporal boundaries $[t_{start},t_{end}]$ and a specific interaction phase, such as start, active, or end. These segments partition the video duration $T$ into manageable chunks, anchoring the memory bank to ensure temporal consistency through complex interaction transitions and topological updates. In each event segment, we flag a keyframe $\ddot{T}_i\in e_i$ with maximum visual quality and minimal occlusion.

\textbf{Spatio-Temporal Relations.} $\mathcal{R}=\{\mathbf{R}_1,\mathbf{R}_2,...,\mathbf{R}_m\}$ describe the interactions occurred within each interaction event segment. Within a specific segment $e_m$, the relation set is defined as $\mathbf{R}_m=\{r_1,r_2,...,r_k\}$, with each individual relation $r_i$ defined as a tuple $r_i = \langle (o_a, o_b), \mathcal{I}, \mathcal{O}_{depth}, \mathcal{S} \rangle$, involving 4 facets:
\begin{itemize}
    \item \emph{Interacting Object Pair} $(o_a, o_b) \in \mathcal{O} \times \mathcal{O}$ defines the pair of entities that carries the interaction.
    \item \emph{Interaction Semantics} $\mathcal{I} = \langle p, c, d \rangle$ defines the semantic predicate $p$ of the interaction (e.g., \textit{cutting}), accompanied by a confidence score $c \in [0, 1]$ and a distance hint $d$ to guide spatial initialization.
    Specifically, the semantic predicates $p$ represents dominant interactions between a pair of entities $(o_a, o_b)$. We use fine-grained atomic verbs \eg, 
\textit{$^{<o_a>}$knife -- $^{<p>}$\textbf{cuts} --$^{<o_b>}$apple}.
    \item \emph{Depth Ordering} $\mathcal{O}_{depth} \in \{\prec_{depth}, \succ_{depth}, \emptyset\}$ establishes the occlusion relation between entities, where $o_a \prec_{depth} o_b$ indicates that $o_a$ consistently occludes $o_b$ over current interaction segment.
    \item \emph{Relative Position} $\mathcal{S} \in \mathcal{H} \times \mathcal{V} \times \mathcal{D}$ provides categorical alignment constraints across the horizontal (left, right, overlap), vertical (above, below, overlap), and depth (front, behind) axes.
\end{itemize}

\subsection{4D Generation and Propagation with HAT-4D}
\label{sec:hat4d}

With the IKG established as a formal constraint engine, HAT-4D orchestrates a multi-agent system to execute the reconstruction. The system transitions from static 3D lifting to dynamic 4D propagation by invoking 3D generation, composition, and 4D propagation skills, with automatic evaluation and rollback to mitigate error accumulation.

\textbf{3D Object Generation and Composition.}
\label{sec:hat4d:3d-comp}
The generation process begins with the \textbf{3D Object Generation Skill}, which initializes the scene at both the first frame and at keyframes $\ddot{T}$ flagged by IKG.
For each frame, the agent first produces a set of virtual anchors for each entity identified by the IKG and then invokes SAM3D~\cite{sam3d} to localize and reconstruct these individual entities as 3D Gaussian Splats.
Since heavy occlusions may cause multiple anchors to be assigned to a single physical entity, the agent leverages object detection to identify and consolidate such instances. For 3D objects reconstructed from multiple anchors belonging to the same physical entity, the agent selectively retains the candidate that most closely aligns with the object semantics specified in the IKG and exhibits the highest reconstruction quality.

Following reconstruction, the \textbf{3D Object Composition Skill} integrates the independent assets into a coherent 3D scene. Guided by the relative spatial timelines defined in the IKG, the agent leverages a lightweight pose optimization operator to refine the 6DoF positions and orientations of each entity. Furthermore, the agent computes exact contact points between entities, if applicable, to ensure the resulting composition is physically plausible and interaction-consistent. Finally, these composed 3D results from the first frame and selected keyframes are cached in a \textbf{memory bank}, serving as robust spatial anchors for subsequent 4D propagation.

\textbf{Memory-Augmented 4D Propagation.}
\label{sec:hat4d:4d-propg}
To extend the composed 3D scene across time, HAT-4D employs a segment-wise \textbf{4D Propagation Operator} (L4GM~\cite{l4gm}), where we render $4$ orthogonal planes from the 3D composition result as spatial initialization. 
The 4D propagation operator is conditioned on both immediately preceding refined frames from the first frame and keyframes in the memory bank, ensuring long-term temporal stability and maintaining strict object identity consistency throughout highly dynamic interaction phases.

\textbf{Multi-Dimensional Evaluation and Rollback.}
\label{sec:hat4d:rollback}
To systematically prevent error accumulation, a dedicated 4D Generation Evaluation Skill serves as the system's critic. For each propagated 4D segment, the agent renders recent frames into multi-view videos and performs a comprehensive assessment driven by the IKG constraints across two dimensions: 
\begin{itemize}
    \item \textit{Dynamic Assessment} evaluates physical plausibility (e.g., interpenetration violations) and long-term memory stability over the temporal sequence;
    \item \textit{Static Assessment} examines visual fidelity, texture quality, and cross-view consistency of the individual assets.
\end{itemize}

When the agent identifies an error, it provides detailed diagnostic descriptions to the corresponding generative agents and triggers a selective rollback. Errors related to dynamic physics trigger a localized re-generation at the 4D propagation stage, whereas artifacts related to static visual quality lead to a complete rollback to the 3D object generation stage for geometric refinement.

\subsection{Multi-Level Human-in-the-Loop Collaboration}
\label{sec:hitl}
Although vision–language models (VLMs) exhibit strong generalization capabilities in understanding physical knowledge under open-world settings, monocular video-based 4D generation remains a highly ill-posed problem. In ambiguous scenarios where multiple plausible interpretations exist, human knowledge is still essential. In this subsection, we describe how human knowledge is incorporated at different stages of the generation pipeline through multi-level refinement operators, guiding VLMs toward more reliable dynamic 3D generation.

\textbf{Online Fine-tuning with Human Feedback.}
During the generation of specific video segments, when the output produced by VLM-driven agents is unsatisfactory, human users can directly correct the generated results. These corrections are then formulated as system prompts and provided to the corresponding agents in subsequent invocations. In this way, human knowledge is injected into an online fine-tuning agent, enabling the system to adapt its behavior with minimal additional cost. The specific prompt formats used by different agents are detailed in the appendix.

\textbf{Multi-level Human Refinement Operators.}
To support efficient human-guided refinement of 3D generation results, HAT-4D provides a set of multi-level refinement operators.
\begin{itemize}
    \item \emph{Gaussian-level Refinement:} Users can directly modify attributes of selected Gaussian Splats, including position, orientation, color, and opacity.
    \item \emph{Region-level Refinement:} Users can select regions with poor generation quality, where local re-generation and optimization are performed using multi-view video generation models via local latent denoising.
    \item \emph{Object-level Refinement:} Specific objects can be re-generated using SAM3D, followed by pose adjustment to better align with the interaction context.
\end{itemize}

Beyond interactive editing, HAT-4D leverages these human-refined outputs to establish a continuous self-improving data engine. The generated 4D assets through the collaboration process serve as high-fidelity pseudo ground-truth for the offline fine-tuning of the underlying learnable 4D generation operators, establishing a scalable bridge for producing robust physical priors in real-world Embodied AI applications.

\begin{figure*}[!t]
    \centering
     \begin{subfigure}{\textwidth} 
        \centering
        \includegraphics[width=\linewidth]{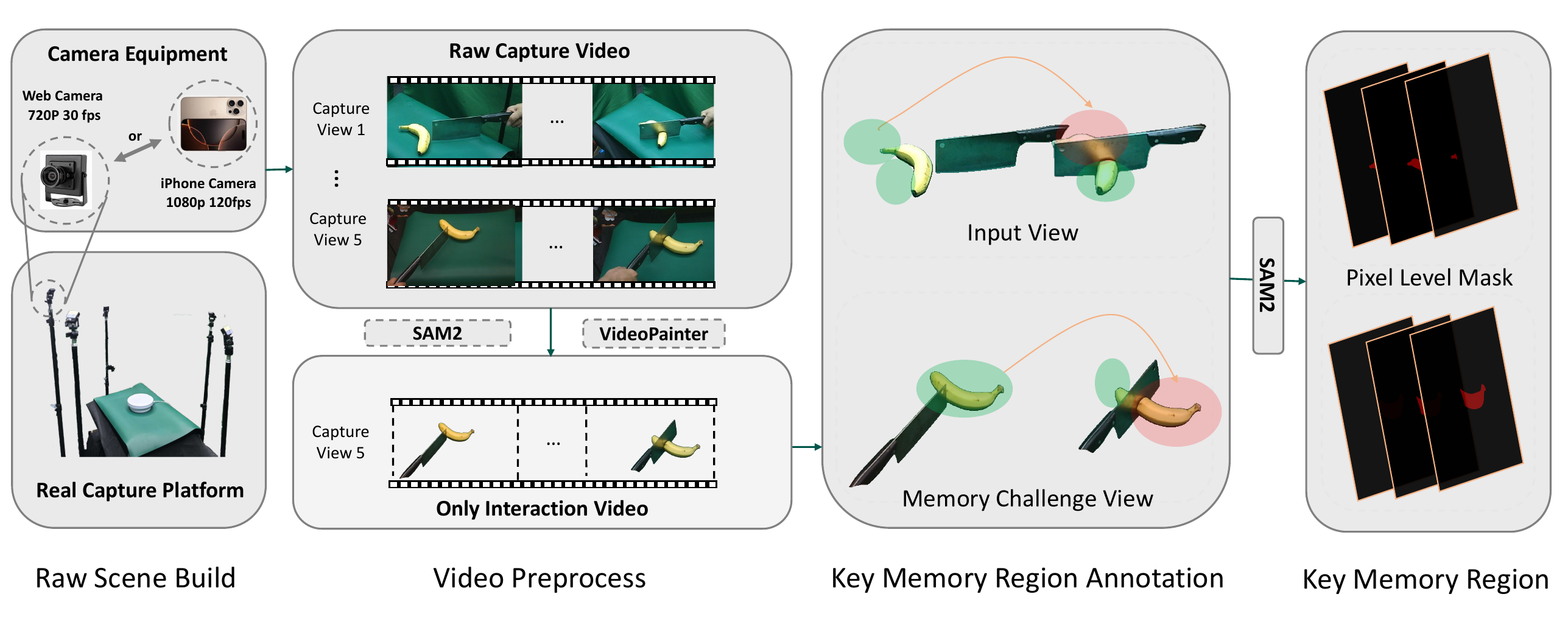} 
       \end{subfigure}
    \caption{
\textbf{Construction of MVOIK-4D and annotation of MemoryOIK-4D.}
We built a multi-view capture platform to record diverse 4D object interaction scenes.
SAM-2 and VideoPainter are used to segment interacting objects and complete occlusions caused by hand manipulation.
For memory scenarios, we select viewpoints where object visibility changes during interaction and annotate the corresponding regions with pixel-level masks to define key memory regions.
}\label{fig：benchmark_pipeline}
\end{figure*}

\section{Benchmark and Metrics}

To systematically evaluate the performance of our proposed HAT-4D framework in reconstructing complex, dynamic object interactions, we introduce a novel benchmark dataset \textbf{MVOIK-4D}, alongside a tailored multi-dimensional evaluation protocol involving overall visual fidelity, spatio-temporal memory retention, and the physical plausibility of the reconstructed interactions.

\subsection{MVOIK-4D: Multi-View Object Interaction Knowledge Dataset}
The Multi-View Object Interaction Knowledge dataset (MVOIK-4D) is constructed using animated 3D assets and a custom multi-camera capture system, with the following two complementary subsets:
\begin{itemize}
    \item \textbf{ToolOIK-4D} focuses on tool-based interactions involving complex deformations. It covers diverse tools (e.g., knife, tongs, lighter) and targets (rigid solids, deformable solids, liquids), featuring 6 highly dynamic interaction categories like liquid splashing, fruit slicing, and peeling.
    
    \item \textbf{MemoryOIK-4D} targets spatio-temporal reasoning in scenarios with dynamic mutual occlusions. It includes \textit{shell games}, \textit{content-revealing interactions under viewpoint changes}, and \textit{tool-based interactions under occlusions}, with few-shot annotations from specific input viewpoints, along with regions of interest that are critical to memory-dependent reconstruction.
\end{itemize}

As in Fig.~\ref{fig：benchmark_pipeline}, we built a multi-camera capture system to acquire real-world object interaction data with calibrated multi-view observations (see Appendix for details). 
For the \textbf{MemoryOIK-4D} subset specifically, we identify input viewpoints where object visibility undergoes deliberate changes across views, and leverage video segmentation annotation tools to label reconstruction regions that are closely related to memory-dependent reasoning.

\subsection{Multi-Dimensional Evaluation}
Standard 3D/4D generation metrics are insufficient for capturing the physical nuances of interacting entities. Therefore, we propose a multi-dimensional evaluation protocol tailored for dynamic multi-object scenarios.

\textbf{Overall Generation Quality.} Following Consistent4D~\cite{consistent4d}, we evaluate the baseline generation quality and temporal consistency by computing CLIP, LPIPS, and FVD scores between the ground-truth multi-view videos captured by multi cameras and the corresponding rendered outputs generated by the models.

\textbf{Temporal Memory Quality.} We evaluate the model's temporal memory capability across three distinct levels:
\begin{itemize}
    \item \textit{Frame-to-Frame Continuity:} We adopt the optical-flow-aligned residual metric from VBench~\cite{huang2023vbenchcomprehensivebenchmarksuite} to measure high-frequency temporal jitter between consecutive frames. We calculate the gradient difference between the ground-truth frame $I_t$ and the warped previous frame $\hat{I_t}$, call it intra-quality. Unlike MSE, which averages out pixel intensities, this metric is sensitive to texture flickering and fine-grained artifacts that are often perceptually disturbing but numerically small in standard metrics.
    \item \textit{Long-Term Memory Stability:} 
    To evaluate the long-term object permanence, we utilize DINOv3~\cite{simeoni2025dinov3} to split one picture into square "patches" and extract the features of every patch. Then we evaluate the similarity of two different patches by cosine-similarity. We define memory patches as "those patches similar to some patch appearing in the past of input, but not visible at present". The final score can be calculated as the ratio of the number of memory patches between the predicted picture and the gt picture.
The detailed analysis of this metric is provided in the appendix.

\end{itemize}

\textbf{Interaction Reconstruction Quality.} We utilize Qwen3-VL\cite{bai2025qwen3vltechnicalreport} to qualitatively rank physical plausibility based on the realism of object deformations and the accuracy of spatial bounds during interaction (e.g., if interpenetration occurred or not).
To further validate the reliability of the Qwen3-VL evaluation, we conduct a user study comparing human judgments with the model's rankings, demonstrating strong consistency between them. 
The detailed prompt design for Qwen3-VL and the correlation analysis with human evaluations are provided in the appendix.

\section{Experiment}

\subsection{Setting}

\paragraph{Baselines.}
To analyze the capability of existing 4D generation models in handling dynamic object interactions, we select several representative end-to-end 4D generation methods as baselines. These models include L4GM~\cite{l4gm}, GVFDiffusion~\cite{gvfdiffusion}, SV4D~\cite{sv4d}, as well as SDS-based approaches such as STAG4D\cite{stag4d} and FB4D\cite{fb4d}.
We adopt the open-source Qwen-VL 235B-A22B-Instruct model as the vision–language model (VLM), SAM3D and L4GM~\cite{l4gm} as 3D/4D generation operators respectively. We use $lr=1\times10^{-5}$ for the pose optimizer, and set the memory bank size to 8, where the model generates 3D Gaussians for the subsequent 7 frames conditioned on the current frame.
As ablation, HAT-4D (w.o. IKG) generates only a simple description of the interaction scene instead of a highly structured and informative IKG. HAT-4D (w.o. memory) limit the size of memory bank is zero.

\subsection{Comparative Experiment}

\begin{table*}[t]
\centering
\renewcommand{\arraystretch}{1.25}
\setlength{\tabcolsep}{3mm}
\caption{\textbf{Quantitative comparison on the MVOIK-4D benchmark.}
Overall generation quality, interaction reconstruction quality, and temporal memory quality are evaluated using decomposed metrics.
\textit{Ideal} indicates the reference optimal value.}
\label{tab:oik4d_main_results}
\resizebox{\linewidth}{!}{
\begin{tabular}{lccccccc}
\toprule
\multirow{2}{*}{\textbf{Method}} 
& \multicolumn{3}{c}{\textbf{Generation Quality}} 
& \multicolumn{2}{c}{\textbf{Interaction Quality}} 
& \multicolumn{2}{c}{\textbf{Memory Quality}} \\
\cmidrule(lr){2-4} \cmidrule(lr){5-6} \cmidrule(lr){7-8}

& \textbf{CLIP} $\uparrow$ 
& \textbf{LPIPS} $\downarrow$ 
& \textbf{FVD} $\downarrow$ 
& \textbf{Deform} $\uparrow$ 
& \textbf{Relation} $\uparrow$ 
& \textbf{Intra} $\downarrow$ 
& \textbf{Long} $\uparrow$ \\

\midrule
\textit{Ideal} 
& 1.0 & 0.0 & 0.0 & 10 & 10 & 0.0 & 1.0 \\
\midrule

L4GM~\cite{l4gm}                 
& 0.8259 & 0.1968 & 962.8593 & 2.5268 & 1.8973 & 0.0013 & 0.0598 \\

GVFDiffusion~\cite{gvfdiffusion} 
& 0.8059 & 0.1961 & 1006.0087 & 2.5871 & 1.8549 & 0.0017 & 0.0494 \\

SV4D~\cite{sv4d}                 
& 0.7914 & 0.1933 & 1362.2214 & 2.4732 & 2.0536 & 0.0017 & 0.0412 \\

STAG4D~\cite{stag4d}                        
& 0.8167 & 0.1674 & 1013.8370 & 2.0915 & 1.5692 & 0.0011 & 0.0650 \\

FB4D~\cite{fb4d}              
& 0.8307 & 0.1576 & 889.7666 & 2.3326 & 1.8929 & 0.0011 & 0.0468 \\

\midrule

HAT-4D (wo IKG)         
& \textbf{0.8370 }&  \underline{0.1435} & \textbf{822.5620} & 3.2835 & \underline{2.4576} & \underline{0.0006} & \underline{0.0708} \\

HAT-4D (wo memory)        
& \underline{0.8319} & 0.1467 & 834.1219 & \underline{3.4213} & 2.5579 & \underline{0.0006} & 0.0621 \\

\textbf{HAT-4D (Ours)}            
& 0.8275 & \textbf{0.1433} & \underline{830.4060} & \textbf{3.4487} & \textbf{2.5737} & \textbf{0.0005} & \textbf{0.0901} \\

\bottomrule

\end{tabular}
}
\end{table*}

Tab.~\ref{tab:oik4d_main_results} reports the performance of HAT-4D, its ablated variants and other monocular video-to-4D baselines on the MVOIK-4D benchmark across multiple evaluation dimensions.

\textbf{Overall quality.}
HAT-4D achieves the best performance on both LPIPS and FVD, indicating significantly improved perceptual quality and temporal realism of the generated 4D sequences. This improvement demonstrates the effectiveness of the agent-driven refinement pipeline. It progressively enhances generation quality through iterative interaction-aware optimization.

\textbf{Interaction quality.}
On the interaction-specific metrics, HAT-4D shows substantial improvements in both deformation accuracy and relational consistency. This suggests that our approach better captures physically plausible object deformations while maintaining correct spatial relationships between interacting objects throughout the dynamic process.

\textbf{Memory consistency.}
HAT-4D achieves state-of-the-art performance across all temporal memory metrics. In particular, our method significantly improves both short-term temporal smoothness and long-horizon consistency. This result demonstrates strong capability in preserving coherent object states across extended temporal sequences and multiple viewpoints.

\textbf{Key Module Ablation.}
Adding IKG improves interaction reconstruction and memory consistency. It increases the Deform, Relation, Intra, and Long scores. However, the longer context from the structured IKG slightly reduces CLIP and FVD performance, likely because it weakens VLM-based discrimination. The memory module mainly improves long-term consistency. It preserves object states and maintains temporal stability over long sequences.

\begin{figure*}[!t]
    \centering
     \begin{subfigure}{\textwidth} 
        \centering
        \includegraphics[width=\linewidth]{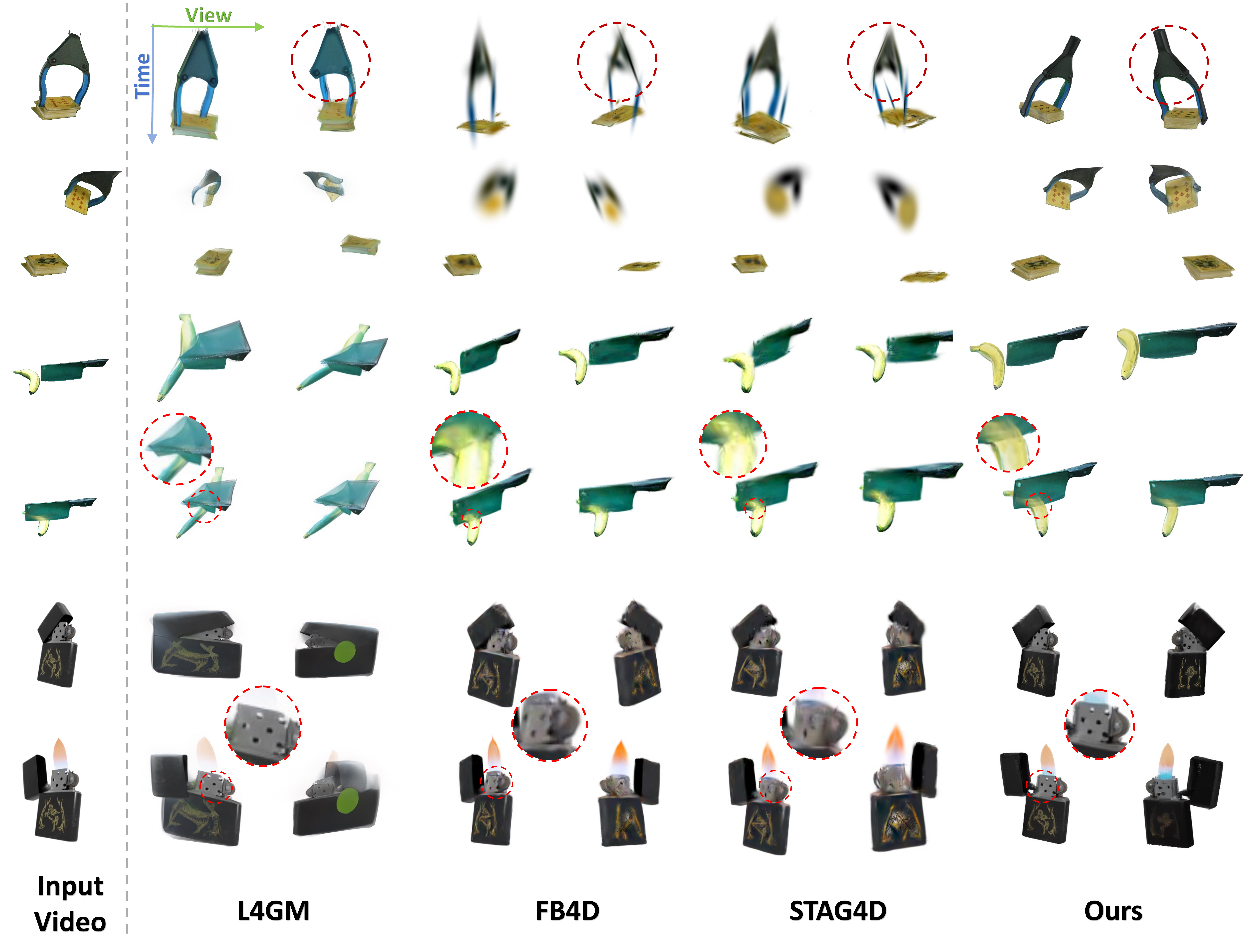} 
       \end{subfigure} 
\caption{\textbf{Qualitative comparison with baseline methods.}
Compared with existing approaches, the agent-driven \textbf{HAT-4D} produces
clearer 3D structures and more stable reconstructions in object interaction regions,
resulting in more consistent dynamic object interaction. } \label{fig：compare_performance}
\end{figure*}

\textbf{Qualitative Analysis}.
As shown in Fig.~\ref{fig：compare_performance}, we present qualitative comparisons between HAT-4D and several baseline methods for monocular reconstruction of object interactions.
The results demonstrate that the progressive evaluation and generation strategy in HAT-4D produces more stable dynamic reconstructions during the generation process.

For example, in the clip–card interaction, the shape of the clip remains clear and consistent across frames, while other methods suffer from structural instability.
Moreover, HAT-4D reconstructs interaction regions more accurately. In challenging cases, such as knife cutting a banana and lighter ignition, our method preserves sharper geometry and clearer interaction boundaries.

In addition, HAT-4D better captures changes in the manipulated objects. For instance, the reconstructed flame exhibits clearer structure and more stable temporal dynamics compared with baseline methods.

\subsection{Ablations about Human Intervention}

\begin{table*}[t]
\centering
\caption{\textbf{Ablation studies on human intervention and refinement operators.}}
\label{tab:ablation_main}

\setcounter{ablationpart}{0}

\footnotesize
\renewcommand{\arraystretch}{1.15}
\setlength{\tabcolsep}{2.0pt}

\begin{adjustbox}{max width=\linewidth,center}
\begin{tabular}{@{}c@{\hspace{4mm}}c@{}}

\begin{tabular}[t]{@{}lccccccc@{}}

\multicolumn{8}{c}{%
    \refstepcounter{ablationpart}%
    \label{tab:ablation_hif}%
    \small
    \textbf{(\alph{ablationpart})}~Human-intervention budget
} \\

\toprule
\multirow{2}{*}{\textbf{HIFs}}
& \multicolumn{3}{c}{\textbf{Generation Quality}}
& \multicolumn{2}{c}{\textbf{Interaction}}
& \multicolumn{2}{c}{\textbf{Memory}} \\

\cmidrule(lr){2-4}
\cmidrule(lr){5-6}
\cmidrule(lr){7-8}

& \textbf{CLIP}$\uparrow$
& \textbf{LPIPS}$\downarrow$
& \textbf{FVD}$\downarrow$
& \textbf{Deform}$\uparrow$
& \textbf{Relation}$\uparrow$
& \textbf{Intra}$\downarrow$
& \textbf{Long}$\uparrow$ \\

\midrule

0
& 0.8341
& 0.1603
& 889.9583
& 3.9038
& 2.8846
& 0.0005
& 0.0012 \\

3
& 0.8489
& \textbf{0.1528}
& 819.5809
& 5.4744
& 4.7885
& 0.0005
& 0.0026 \\

5
& \textbf{0.8521}
& 0.1532
& 799.6682
& \textbf{5.9295}
& \textbf{5.0256}
& 0.0005
& \textbf{0.0027} \\

7
& 0.8518
& 0.1534
& \textbf{798.4898}
& 5.8974
& 4.9423
& 0.0006
& 0.0023 \\

\bottomrule
\end{tabular}

&

\begin{tabular}[t]{@{}lccccccc@{}}

\multicolumn{8}{c}{%
    \refstepcounter{ablationpart}%
    \label{tab:ablation_refinement}%
    \small
    \textbf{(\alph{ablationpart})}~Multi-level refinement operators
} \\

\toprule
\multirow{2}{*}{\textbf{Setting}}
& \multicolumn{3}{c}{\textbf{Generation Quality}}
& \multicolumn{2}{c}{\textbf{Interaction}}
& \multicolumn{2}{c}{\textbf{Memory}} \\

\cmidrule(lr){2-4}
\cmidrule(lr){5-6}
\cmidrule(lr){7-8}

& \textbf{CLIP}$\uparrow$
& \textbf{LPIPS}$\downarrow$
& \textbf{FVD}$\downarrow$
& \textbf{Deform}$\uparrow$
& \textbf{Relation}$\uparrow$
& \textbf{Intra}$\downarrow$
& \textbf{Long}$\uparrow$ \\

\midrule

Agent
& 0.8330
& 0.2126
& 860.5518
& 4.1000
& 3.0250
& 0.0006
& 0.0013 \\

Obj.
& 0.8641
& 0.2060
& 685.7023
& \textbf{6.1750}
& 5.1500
& 0.0006
& 0.0014 \\

Obj.+Reg.
& 0.8657
& 0.2051
& \textbf{681.7343}
& 5.8000
& 5.1500
& 0.0006
& \textbf{0.0016} \\

Obj.+GS
& \textbf{0.8659}
& \textbf{0.2044}
& 686.9764
& 5.8750
& \textbf{5.2500}
& 0.0006
& 0.0014 \\

\bottomrule
\end{tabular}

\end{tabular}
\end{adjustbox}
\end{table*}

\subsubsection{Different Human Intervention.}
We study how different levels of human intervention influence the performance of HAT-4D.
The experiment is conducted on 39 sequences from the MVOIK-4D dataset, among which 17 sequences contain challenging memory-intensive interactions.
During generation, we limit the maximum number of human annotations and evaluate the performance under different human intervention frequencies.

\begin{figure}[!t]
    \centering

    \begin{subfigure}[t]{0.60\linewidth}
        \centering
        \includegraphics[width=\linewidth]
        {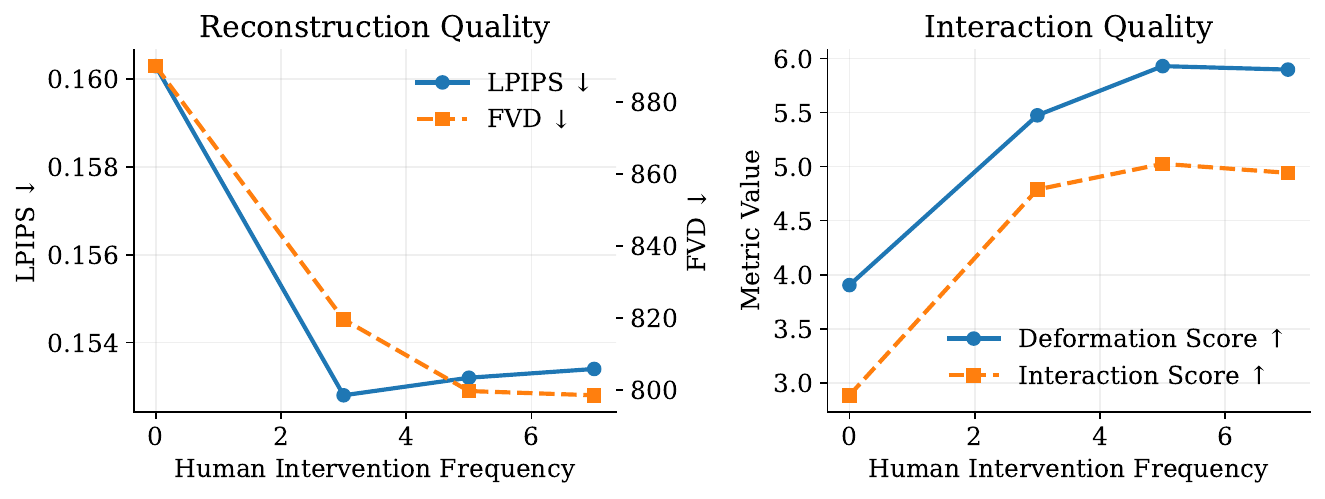}

        \caption{
        \textbf{Impact of human intervention on L4GM finetuning.}
        HIF denotes the number of human annotations allowed during
        generation, with HIF$=0$ representing a fully agent-driven
        pipeline. A small number of human interventions substantially
        improves reconstruction and interaction quality.
        }
        \label{fig:ablation_human_intervention}
    \end{subfigure}
    \hfill
    \begin{subfigure}[t]{0.38\linewidth}
        \centering
        \includegraphics[width=\linewidth]
        {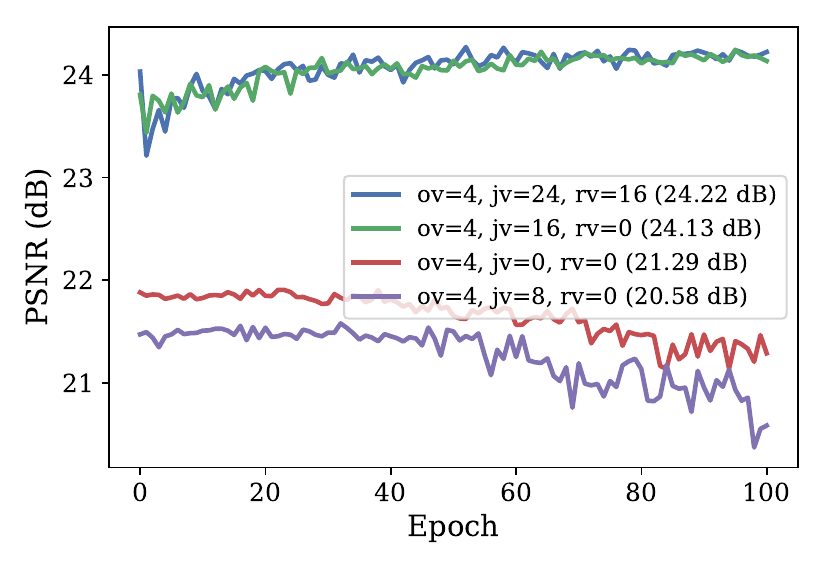}

        \caption{
        \textbf{View scaling.}
        L4GM finetuning under different input-view configurations on
        HAT-4D sequences. \textit{ov}, \textit{jv}, and \textit{rv}
        denote original, jittered, and randomly sampled views,
        respectively.
        }
        \label{fig:psnr_vs_epoch}
    \end{subfigure}

    \caption{
    \textbf{Impact of human intervention and view scaling on
    4D reconstruction.}
    }
    \label{fig:human_view_ablation}
\end{figure}

As shown in Tab.~\ref{tab:ablation_hif} and Fig.~\ref{fig:ablation_human_intervention}, even limited human intervention consistently improves reconstruction quality, interaction accuracy, and temporal consistency over the fully agent-driven setting.

The improvement is particularly significant when only a few interventions are allowed.
Correcting several critical frames is sufficient to fix major errors in the generated dynamics and interaction states.
As the number of interventions increases further, the performance gain gradually saturates.
This observation indicates that sparse human knowledge can effectively guide the generation process and substantially improve the reliability of dynamic 4D reconstruction.

\subsubsection{Different Refinement Operators.}
We evaluate different refinement operators on 10 randomly sampled cases.
For each setting, volunteers can only use the specified refinement tools.
The fully agent-driven pipeline serves as the baseline.

As shown in Tab.~\ref{tab:ablation_refinement}, object-level refinement yields the largest gain by correcting geometry, pose, occlusion, and interaction errors. Region-level and Gaussian-level refinements provide complementary improvements: the former achieves the best FVD and Long scores, while the latter performs best on CLIP, LPIPS, and Relation by reducing local artifacts and appearance inconsistencies.

Overall, object-level refinement is the primary correction tool.
Region-level and Gaussian-level operators are used for fine-grained local adjustments.
This is consistent with our practical annotation workflow.

\subsubsection{Baseline Finetuned in Different Views }

We select 63 high-quality annotated scenes from MVOIK-4D and split them into training and test sets with a ratio of 9:1.
To follow the training protocol of L4GM, each scene is divided into clips with 8 frames.
This produces 1,275 training clips and 97 test clips. For each scene, we choose one real captured viewpoint as the reference view.
Four orthogonal views are rendered around it and used as input views.
Another four captured viewpoints are reserved as validation views for evaluation.

To study the effect of view supervision, we render 24 jittered views from small angular and radial perturbations and 16 uniformly sampled random views. 
Fig.~\ref{fig:human_view_ablation} shows the PSNR curves during the finetuning of L4GM on MVOIK-4D.
With limited supervision views, jittered views alone do not improve the reconstruction quality of L4GM, as the model tends to overfit to a narrow set of viewpoints, which leads to unstable training and lower PSNR.

Increasing the number of supervision views improves the performance of L4GM.
Randomly sampled views provide stronger geometric diversity and better spatial coverage, enabling more stable optimization and higher reconstruction accuracy.
This highlights the importance of  stronger multi-view supervision and  the meaning of HAT-4D for dynamic 4D generation.

\begin{figure*}[!t]
    \centering
     \begin{subfigure}{\textwidth} 
        \centering
        \includegraphics[width=\linewidth]{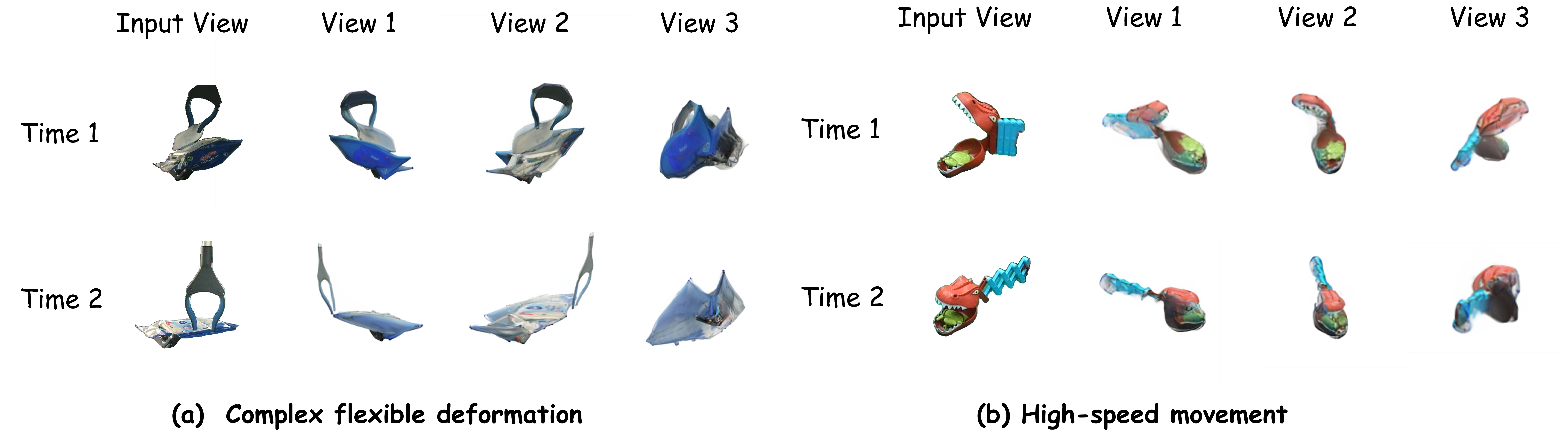}
       \end{subfigure}

\caption{\textbf{Failure cases.}
Left: pressing and folding a thin plastic carton causes complex deformation, distorting novel-view geometry and over-smoothing details.
Right: the spring toy’s rapid extension and recoil cause motion blur and geometric inconsistencies.}\label{fig:failure_cases_in_main}
\end{figure*}

\section{Conclusion}
We present \textbf{HAT-4D}, an agentic framework for reconstructing dynamic 4D object interactions from monocular videos. 
By integrating multi-level human-in-the-loop feedback, HAT-4D refines 3D object generation, spatial composition, and 4D dynamic propagation, improving physical plausibility and temporal consistency. 
We also introduce \textbf{MVOIK-4D}, a benchmark with \textbf{112 scenes}, \textbf{77 tasks}, \textbf{39 interaction categories}, and \textbf{15 object deformation categories}, together with a multi-dimensional evaluation protocol for generation quality, interaction consistency, and long-term memory stability.

However, as shown in Fig.~\ref{fig:failure_cases_in_main}, HAT-4D remains challenged by complex flexible deformation and fast non-rigid motion, where irregular geometry and motion blur degrade SAM3D-based geometry reconstruction. Performance also depends on the reasoning capability and inference efficiency of the underlying VLM. 
Future work will explore improved temporal correspondence, deformation modeling, VLM fine-tuning, and more scalable 4D interaction data generation from monocular videos. 
We hope HAT-4D and MVOIK-4D will support research in object interaction understanding, 4D generation, and robotic perception.

\section*{Acknowledgements}
This work was supported by the Shanghai Municipal Special Program for Basic Research on General AI Foundation Models (Grant No. 2025SHZDZX025G14), National Natural Science Foundation of China (U25A20442, 62306175), Ant Group.

%
%
\bibliographystyle{splncs04}
\bibliography{main}

\end{document}